\documentclass[journal]{IEEEtran}

\usepackage{amsfonts}
\usepackage{amsmath}
\usepackage{amssymb}
\usepackage{mathabx}
\usepackage{mathrsfs}
\usepackage{tensor}
\usepackage{esint}
\usepackage{tikz}
\usepackage{multirow}
\usepackage{makecell}
\usepackage{textcomp}
\usepackage{amsthm}
\usepackage{subfigure}
\usepackage{graphicx}
\usepackage{stackengine}
\usepackage{balance}
\usepackage[caption=false,font=footnotesize,labelfont=sf,textfont=sf]{subfig}
\theoremstyle{remark}

\begin{document}
\title{Coordinate-Based Neural Representation Enabling Zero-Shot Learning for 3D Multiparametric Quantitative MRI}
\author{Guoyan Lao$^{1}$, Ruimin Feng$^{1}$, Haikun Qi$^{2}$, Zhenfeng Lv$^{2}$, Qiangqiang Liu$^{3}$, Chunlei Liu$^{4}$, \\ Yuyao Zhang$^{5}$, and Hongjiang Wei$^{1,*}$
\\ \IEEEauthorblockA{$^{1}$School of Biomedical Engineering, Shanghai Jiao Tong University, Shanghai, China.} 
\\ \IEEEauthorblockA{$^{2}$School of Biomedical Engineering, ShanghaiTech University, Shanghai, China} 
\\ \IEEEauthorblockA{$^{3}$Department of Neurosurgery, Shanghai Jiao Tong University School of Medicine, Shanghai, China}
\\ \IEEEauthorblockA{$^{4}$Department of Electrical Engineering and Computer Sciences, University of California, Berkeley, CA, USA}
\\ \IEEEauthorblockA{$^{5}$School of Information Science and Technology, ShanghaiTech University, Shanghai, China}

\thanks{*Corresponding author (e-mail:hongjiang.wei@sjtu.edu.cn).}
}

\IEEEtitleabstractindextext{

\begin{abstract}
Quantitative magnetic resonance imaging (qMRI) offers tissue-specific physical parameters with significant potential for neuroscience research and clinical practice. However, lengthy scan times for 3D multiparametric qMRI acquisition limit its clinical utility. Here, we propose SUMMIT, an innovative imaging methodology that includes data acquisition and an unsupervised reconstruction for simultaneous multiparametric qMRI. SUMMIT first encodes multiple important quantitative properties into highly undersampled k-space. It further leverages implicit neural representation incorporated with a dedicated physics model to reconstruct the desired multiparametric maps without needing external training datasets. SUMMIT delivers co-registered T1, T2, T2*, and quantitative susceptibility mapping. Extensive simulations and phantom imaging demonstrate SUMMIT's high accuracy. Additionally, the proposed unsupervised approach for qMRI reconstruction also introduces a novel zero-shot learning paradigm for multiparametric imaging applicable to various medical imaging modalities.
\end{abstract}

}
\maketitle
\IEEEdisplaynontitleabstractindextext
\IEEEpeerreviewmaketitle

\section{Introduction}
\label{sec:Introduction}
\IEEEPARstart{Q}{uantitative} magnetic resonance imaging (qMRI) provides tissue-specific quantification of relaxation times ($T_1$, $T_2$, and $T_2^*$), water molecule diffusivity, and tissue magnetic susceptibility. These parameters offer distinct physical insights into brain tissue, allowing for detailed characterization of the local microstructural environment. Increasing evidence supports that multiparametric qMRI facilitates the investigation of tissue characteristics \cite{benjamini2023mapping}, such as quantifying iron, myelination and cell membranes in healthy and diseased brains \cite{vrenken2006whole,kirov2010brain,uchida2019voxel,sawlani2020multiparametric,smits2021mri}. However, multiparametric quantitative mapping typically requires multiple scans using separate MRI sequences, i.e., multi-echo spin echo (ME-SE) for $T_2$ mapping \cite{ben2015rapid}, inversion-recovery spin echo (IR-SE) for $T_1$ mapping \cite{shao2017accuracy}, and multi-echo gradient-recalled echo (ME-GRE) for $T_2^*$ and magnetic susceptibility mapping \cite{feng2018quantitative}. This process can prolong scan times and introduce potential misalignment between images, limiting clinical applicability. Recently, simultaneous multiparametric qMRI (MP-qMRI) has rapidly developed and become a vital tool for enhancing qMRI scan efficiency \cite{wang2019echo,christodoulou2018magnetic,ma2013magnetic,metere2017simultaneous,chen2018strategically,zhang2022blip}. These techniques are specifically designed to obtain co-registered multiple quantitative maps using a single MRI scan, significantly reducing total scan times and eliminating the spatial misalignment between maps. Therefore, simultaneous MP-qMRI holds the potential for exploring interlinks between different quantitative maps and improving diagnosis precision in clinical settings.

Currently, techniques such as magnetic resonance fingerprinting (MRF) \cite{ma2018fast, boyacioglu20213d}, echo planar time-resolved imaging (EPTI) \cite{wang2019echo, wang20223d}, and MR multitasking \cite{christodoulou2018magnetic, cao2022three} have demonstrated the capability to simultaneously generate 3D $T_1$, $T_2$, $T_2^*$, or quantitative susceptibility mapping (QSM) images \cite{li2023apart, shin2021chi, chen2021decompose}, providing a more comprehensive characterization of tissue properties, offering advantages over conventional qualitative MRI in terms of sensitivity and comparability. These techniques capture the signal evolution induced by multiple tissue properties and encode them into the measured frequency domain, i.e., k-space, which is sparsely sampled to accelerate the acquisition. Therefore, the recovery of tissue relaxation and magnetization properties becomes a complex inverse problem. Methods like the dictionary-matching method \cite{ma2013magnetic, bipin2019magnetic} and the low-rank tensor (LRT)-based method \cite{he2016accelerated, zhang2015accelerating} have been proposed to address this challenge. The dictionary-matching method, however, suffers from high computational complexity, making it impractical for 3D simultaneous MP-qMRI, especially with large dictionaries and high-resolution data. The LRT-based method leverages spatiotemporal correlations of signals for dimensionality reduction, enabling reconstruction from highly sparse measurements. While promising in EPTI and MR multitasking, this reconstruction approach involves multiple steps, including relaxation basis estimation, weighted image tensor reconstruction, and Bloch equation fitting. Thus, errors at any of these steps can propagate to the final reconstruction of quantitative maps, affecting quantification accuracy. Additionally, the methods mentioned above cannot simultaneously recover underlying tissue quantitative parameters and coil sensitivities from multi-coil k-space data. They always require pre-calculated coil sensitivity maps derived from prescan, which might introduce errors propagating to the subsequent reconstruction. Consequently, the limitations of current reconstruction methods hinder further improvements in imaging speed and quantification accuracy for simultaneous MP-qMRI.

In recent years, deep networks have demonstrated significant application potential in the field of MR image reconstruction. However, the extensive training data required by supervised deep learning methods limits their application in simultaneous MP-qMRI reconstruction. In contrast, unsupervised deep learning strategies overcome this constraint by reconstructing images directly from undersampled measurements without requiring additional training data. A more recent study, Joint-MAPLE explored an unsupervised convolutional neural networks (CNN) for simultaneous MP-qMRI reconstruction \cite{heydari2024joint}. This method integrates CNNs with the unrolled physical model to learn a series of weighted images, while incorporating joint optimization of quantitative parameter maps. Although this process avoids the issue of error propagation, Joint-MAPLE provides a limited number of qMRI parametric images, and its further application to an increasing number of parameters could be hindered by the substantial computational cost associated with 3D CNNs. Additionally, the spectral bias towards low-frequency information limits the ability of CNN-based structures to capture subtle image details. Recently, implicit neural representation (INR) has emerged as a new paradigm for unsupervised learning, achieving remarkable results in medical image reconstruction \cite{shen2022nerp, feng2023imjense}. INR trains a multi-layer perceptron (MLP) to represent the medical image as a continuous function that maps the spatial coordinates to the image intensities. Specifically, INR utilizes various spatial coordinate embedding techniques \cite{sitzmann2020implicit, tancik2020fourier, wu2023self} to enhance its capability to capture high-frequency image details. Technically, INR exceeds CNN-based structures in two key aspects: 1) it alleviates the low-frequency bias issue of CNNs, and 2) it provides a spatially continuous representation of the reconstructed image, which has been proven more practical for incorporating various imaging forward models. Consequently, INR naturally excels in solving complex inverse problems for medical imaging \cite{feng2023imjense, wu2023self, wu2022arbitrary}, offering the potential to address the challenges currently in simultaneous MP-qMRI reconstruction.

In this study, we introduce an innovative imaging methodology for 3D SimUltaneous MultiparaMetric quantitative MRI via Implicit neural represenTation, referred to as SUMMIT.  SUMMIT encompasses a well-designed data acquisition process comparable in time to typical clinical scans, and an innovative zero-shot MP-qMRI reconstruction based on INR. Our approach encodes multiple important quantitative tissue properties into highly undersampled multi-dimensional k-space measurements. Specifically, the proposed imaging sequence employs $T_2$-prepared inversion recovery and ME-GRE readout modules to encode images at different inversion and echo times with different $T_2$ weightings into k-space.

Furthermore, SUMMIT leverages INR to solve the ill-posed inverse problem relating the subsampled k-space to the quantitative maps of the magnetic parameters. It models the underlying tissue quantitative parameter maps and coil sensitivities as continuous functions of spatial coordinates and decodes the parameter maps directly from undersampled k-space measurements in an unsupervised manner. Under this representation, SUMMIT reformulates the multi-step complicated inverse problem of simultaneous MP-qMRI into a single-step function parameter optimization problem, guided by the dedicated physics model. Benefiting from the powerful fitting capability of neural networks, our method achieves single-step reconstruction of co-registered $T_1$, $T_2$, $T_2^*$, and QSM mappings from the highly undersampled k-space data.

Finally, SUMMIT delivers six different high-resolution quantitative MR images within a well-controlled acquisition time. To the best of our knowledge, SUMMIT is the most comprehensive qMRI reconstruction from a single data acquisition. The results of the simulation and phantom data demonstrate the superiority of our proposed method compared to existing methods, significantly reducing the quantification errors by 11.0\%, 11.8\%, and 4.8\% on $T_1$, $T_2$, and $T_2^*$ maps at high acceleration factor. We thus believe that SUMMIT could detect essential and complex changes across many qMRI parameters simultaneously, which are crucial for computer-aided MR analysis.

\section{Methods}
\label{sec:Methods}

\begin{figure*}[ht]
    \centering
    \includegraphics[width=160mm]{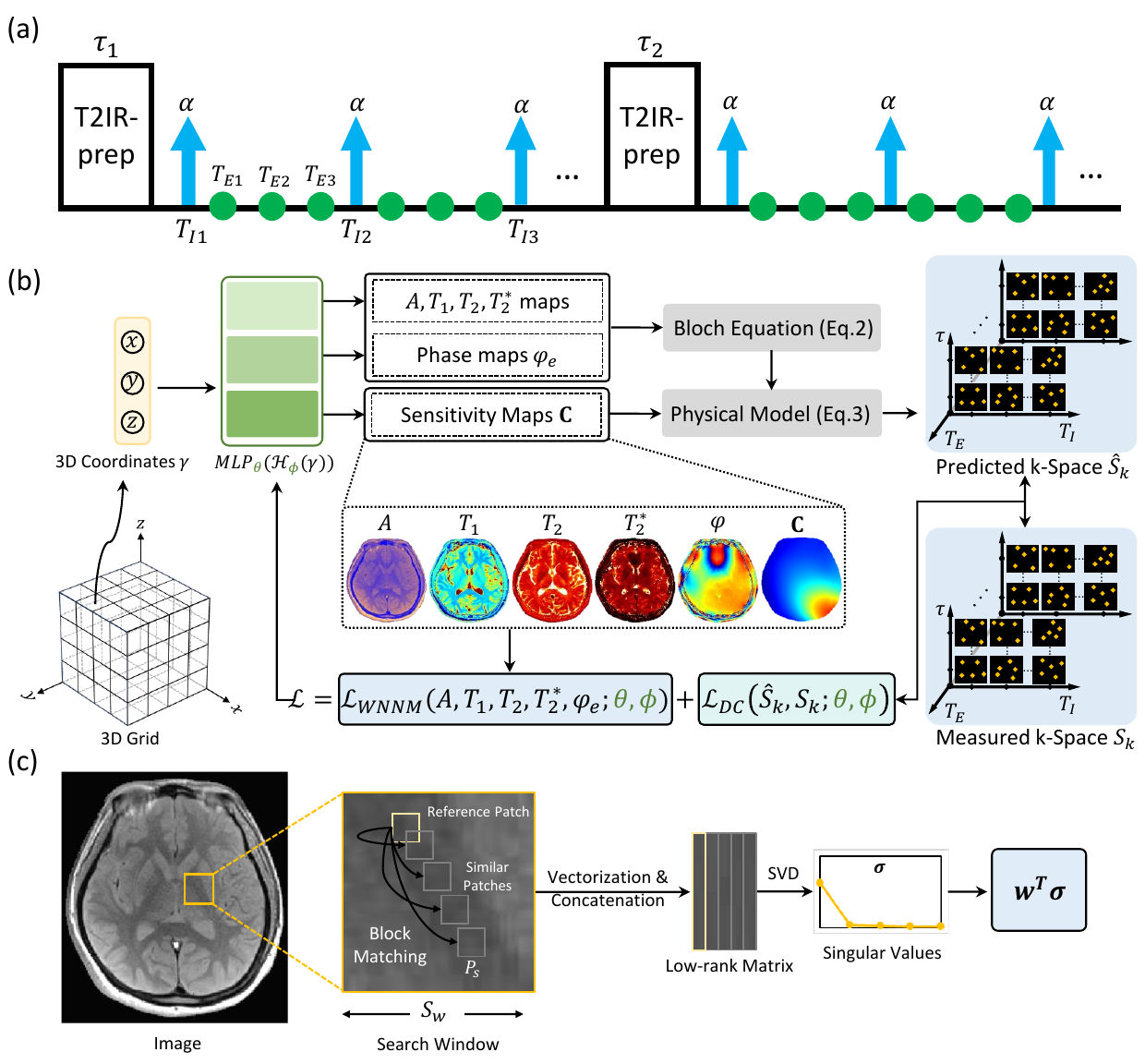}
    \caption{
    Overview of SUMMIT. (a) Diagram of the MRI sequence for data acquisition. (b) The reconstruction process of SUMMIT. The 3D coordinates $\gamma$ are fed into the encoding module $H_\phi$ and $MLP_\theta$ to simultaneously estimate the multiparametric quantitative maps and sensitivity maps. The predicted multiple-dimensional k-space $\hat{S}_{k}$ is generated through the Bloch equation and the physical forward model based on the estimated maps. The weights in the MLP and encoding module are optimized by minimizing the data consistency (DC) loss and weighted nuclear norm minimization (WNNM) loss. (c) The denoising procedure of WNNM module. First, similar patches are extracted through block matching. These patches are then vectorized and concatenated along one dimension to construct a low-rank matrix. Finally, singular value decomposition (SVD) is performed to obtain a series of singular values $\sigma$. The WNNM loss is a weighted sum of the singular values. 
    }
    \label{fig1}
\end{figure*}

\subsection{MRI Sequence and Data Acquisition}
As illustrated in Fig. \ref{fig1}(a), the MRI sequence consists of $T_2$-prepared inversion recovery (T2IR-prep) modules and ME-GRE readout (green circle) with fast low angle shot (FLASH) excitations (blue arrows). $T_2$ contrasts are generated by varying T2IR-prep durations $\tau$, $T_1$ contrasts are determined by the inversion times $T_I$, and $T_2^*$ contrasts are produced by different echo times $T_E$ of the ME-GRE readout. This setup allows us to represent k-space as a multiple-dimensional tensor $S_k (k_x,k_y,k_z,T_I,\tau,T_E)$ with three spatial dimensions $(k_x,k_y,k_z)$ and three temporal dimensions $(T_I,\tau,T_E)$. Fully sampling this high-dimensional k-space is time-consuming, so a highly undersampled strategy is necessary for accelerating the scan. To enhance sampling incoherence, we adopt a random Cartesian sampling pattern with a variable density Gaussian distribution along phase encoding directions ($k_y$ and $k_z$). The acceleration factor is defined as:
\begin{equation}
    R = \frac{n_{k_{y}}*n_{k_{z}}}{n_{segment}*n_{sampled}},
\label{eq1}
\end{equation} 
where $n_{k_{y}}$ and $n_{k_{z}}$ are the steps of phase encoding, $n_{segment}$ represents the number of segments (FLASH excitations) within one inversion recovery period, and $n_{sampled}$ is the number of sampled k-space points in the $k_y$-$k_z$ plane per segment.

The signal intensity at a given pixel is governed by the $T_1$, $T_2$, and $T_2^*$ relaxations of the tissue at that location, which are described by the following Bloch equation: 
\begin{equation}
\begin{split}
    S = A\frac{1 - e^{- \frac{T_{R}}{T_{1}}}}{1 - e^{- \frac{T_{R}}{T_{1}}}cos\alpha}sin\alpha \cdot e^{- \frac{T_{E}}{T_{2}^{*}}} \cdot e^{j\varphi_{e}} \cdot \\ 
    \left\lbrack 1 + ( {Be^{- \frac{\tau}{T_{2}}} - 1})({e^{- \frac{T_{R}}{T_{1}}}cos\alpha})^{n}\right\rbrack,
\label{eq2}
\end{split}
\end{equation}
where $A$ is the equilibrium amplitude, $B$ represents the inversion efficiency, $T_R$ is the repetition time, $T_E$ is the echo time, $\tau$ is the T2IR-prep duration, $\alpha$ is the flip angle, $\varphi_e$ denotes the phase at echo $e$, and $n$ is the index of the segment within one inversion recovery period. The undersampled k-space signal $S_k$ is described by the physical forward model:
\begin{equation}
    S_{k} = \mathbf{M}\mathbf{F}\mathbf{C}S,
\label{eq3}
\end{equation} 
where $\mathbf{C}$ are the coil sensitivity maps, $\mathbf{F}$ represents the Fourier transform and $\mathbf{M}$ denotes the sampling mask. Therefore, the k-space can be represented as a six-dimensional tensor $S_k (k_x,k_y,k_z,T_I,\tau,T_E)$ with tissue quantitative parameters encoded.

\subsection{Continuous Representation of Tissue Parameters}
In this study, we model the tissue parameters as continuous functions of spatial coordinates, with these functions taking the 3D spatial coordinate $\gamma=(x,y,z)$ as input and yielding the corresponding values of $A$, $T_1$, $T_2$, $T_2^*$, $\varphi_e$ and $\mathbf{C}$ at that location. Let $\vartheta$ represent the parameters in these continuous functions. Then, the signal in the image and k-space domain can be parameterized as $S(\vartheta)$ and $S_k(\vartheta)$, respectively. Under this continuous function representation, the recovery of potential tissue parameters can be formulated as:
\begin{equation}
    \underset{\vartheta}{arg~min}~{g(S_{k}(\vartheta) - \mathbf{M}\mathbf{F}\mathbf{C}S(\vartheta)) + \lambda\mathcal{R}( {S(\vartheta)})},
\label{eq5}
\end{equation} 
where $g(S_k(\vartheta)-\mathbf{M}\mathbf{F}\mathbf{C}S(\vartheta))$ denotes the data consistency term and $\mathcal{R}(S(\vartheta))$ represents the regularization term. Thus, the traditional complicated nonlinear inverse problem has been transformed into the continuous function parameter optimization problem by searching for the optimal solution in the parameter domain. Once these function parameters are optimized, we can obtain the desired quantitative maps by inputting the corresponding coordinates.

We adopt the deep neural network to implicitly represent these continuous functions, which consists of a parameterized encoding module $H_\phi$ and a multilayer perceptron $MLP_\theta$. Here, $\phi$ and $\theta$ respectively denote the weights in the encoding module and MLP that need to be optimized. The weights $\phi$ and $\theta$ are iteratively optimized by minimizing the following loss function:
\begin{equation}
    \mathcal{L} = \mathcal{L}_{DC}({{\hat{S}}_{k},S_{k};\theta,\phi}) + \lambda\mathcal{L}_{WNNM}(A,T_{1},T_{2},T_{2}^{*};\theta,\phi),
\label{eq4}
\end{equation} 
where $\mathcal{L}_{DC}$ imposes the data consistency between the predicted k-space signal $\hat{S}_{k}$ and the measured k-space signal $S_{k}$, $\mathcal{L}_{WNNM}$ represents the weighted nuclear norm minimization (WNNM) \cite{gu2014weighted} regularization term for joint denoising multiple quantitative maps and $\lambda$ balances the contributions of these two terms.

\subsection{MLP with Hash Encoding}
Considering that MLPs are universal function approximators, we employ MLPs with hash encoding \cite{muller2022instant} to learn these continuous functions. The hash encoding maps the 3D coordinate $\gamma$ into a higher dimension, which is then fed into the MLP. This process enhances the capability to fit high-frequency information. Compared to other encoding methods, hash encoding provides a more compact signal representation, incorporating a significant portion of trainable parameters, thus allowing the use of smaller MLPs. Consequently, the combination of MLP and hash encoding significantly reduces GPU memory consumption, making the proposed method suitable for handling multiple-dimensional MP-qMRI data. Specifically, in hash encoding, the trainable encoding parameters are organized into $L$ different resolution levels, represented as $L$ independent hash tables. These hash tables express a series of resolutions that increase geometrically, with the coarsest resolution denoted as $N_{min}$ and a growth factor denoted as $b$. Therefore, these resolutions are $N_{min}$, $b{\times}N_{min}$, ..., $b^{L-1}{\times}N_{min}$. Each resolution level contains $T$ feature vectors, and the dimensionality of each feature vector is $F$.

We stabilize and constrain the learning of continuous functions by adjusting the configurations of hash encoding and MLPs based on the characteristics of the target image, thereby narrowing down the solution space of the optimization problem and yielding improved outcomes. Specifically, the activation function of the last layer in MLPs is an exponential function for $A$, $T_1$, $T_2$, and $T_2^*$ maps, ensuring that the outputs are all positive. Since the phase accumulation has both positive and negative values, we use separate MLPs without activation functions at the last layer. The hyperparameters for the hash encoding of the quantitative maps are set to their default values. Coil sensitivity maps inherently exhibit smooth characteristics and lie in a low-dimensional space, thereby we control the representation of coil sensitivity by reducing the resolution and the total number of parameters of the hash encoding. These configurations of hash encoding and MLP primarily consider the characteristics of different quantitative maps, contributing to the stability of the optimization process and being key factors for the network's convergence. These settings are kept consistent across all datasets.

\subsection{Weighted Nuclear Norm Minimization Loss Function}
Nuclear norm minimization is a general category for low-rank matrix approximation methods, which can be used to design denoising algorithms \cite{ji2010robust, wang2013nonlocal, gu2014weighted}. To improve the flexibility of nuclear norm and leverage priors on the matrix singular values, Gu \textit{et al.} proposed the following weighted nuclear norm minimization (WNNM) problem: 
\begin{equation}
    \min\limits_{\mathbf{X}}{\left\|{\mathbf{Y}-\mathbf{X}}\right\|_{F}^{2} + \left\|\mathbf{X}\right\|_{\mathbf{w},*}},
\label{eq6}
\end{equation}
where \textbf{Y} is the given matrix and WNNM aims to approximate \textbf{Y} by \textbf{X} while minimizing the weighted nuclear norm of matrix \textbf{X} defined as:
\begin{equation}
    \left\|\mathbf{X}\right\|_{\mathbf{w},*}={\sum\limits_{i}\left|{w_{i}\sigma_{i}(\mathbf{X})}\right|_{1}},
\label{eq7}
\end{equation}
where $\sigma_{i}(\mathbf{X})$ denotes the $i^{th}$ singular value of $\mathbf{X}$, $\mathbf{w}=\left\lbrack{w_{1},w_{2},\mathbf{~}\ldots,w_{n}}\right\rbrack^{T}$ and $w_{i}$ is a non-negative weight assigned to $\sigma_{i}(\mathbf{X})$.

To effectively guide the optimization process of the hash encoding and MLPs, we apply WNNM as an explicit regularization term on the $A$, $T_1$, $T_2$, $T_2^*$ and $\varphi_e$ maps generated by the MLPs. Compared to conventional total variation, WNNM leverages the non-local structural similarity of images, better preserving detailed tissue information while suppressing noise and artifacts. The computation process of WNNM is illustrated in Fig. \ref{fig1}(c). First, block matching is performed. Specifically, for each reference image patch (with a patch size of $P_s$), the $N_p$ most similar patches are matched within a search window of size $S_w$ by calculating the Euclidean distance. These patches are then vectorized and concatenated along one dimension to construct a low-rank matrix. Finally, singular value decomposition (SVD) is used to obtain a series of singular values $\boldsymbol{\sigma}=\left\lbrack {\sigma_{1},\sigma_{2},\mathbf{~}\ldots,\sigma_{N_{p}}}\right\rbrack^{T}$. The larger singular values represent the energy of the main components of the low-rank matrix, i.e., the information of the image itself, while the smaller singular values mainly represent noise and artifacts. Therefore, by assigning smaller weights to larger singular values and larger weights to smaller singular values, noise, and artifacts can be suppressed while effectively preserving detailed tissue information. The calculation of the weight corresponding to the $i^{th}$ singular value $\sigma_i$ is as follows:
\begin{equation}
    w_{i} = \frac{1}{\sqrt{\max\left( {\sigma_{i}^{2} - N_{p}\sigma_{noise}^{2},0} \right)} + \epsilon},
\label{eq8}
\end{equation}
where $\epsilon$ = $10^{-16}$ prevents the denominator from being zero, and $\sigma_{noise}^2$ denotes the variance of the noise, which is approximated by the variance of the residual image. The residual image is obtained by subtracting the mean of the similar patches from the reference patch.

\subsection{Implementations}
The MLP in this study consists of three hidden layers, each containing 64 neurons followed by the ReLU activation function. The hash encoding and MLPs are optimized using the Adam algorithm \cite{kingma2014adam}, with an initial learning rate set to 1$\times$10$^{-3}$, which decays to 0.5 times its previous value every 20 epochs. The total number of epochs is set to 100. For the calculation of WNNM, the parameters are set as $P_s$ = 5, $S_w$ = 7, and $N_p$ = 3. The WNNM loss function is applied simultaneously to the network outputs $A$, $T_1$, $T_2$, $T_2^*$ and $\varphi_e$, with different $\lambda$ values assigned according to the range of each quantitative map:
\begin{equation}
\begin{split}
    \mathcal{L}_{WNNM} = \lambda_0\mathbf{w}_{A}^{T}\mathbf{\sigma}_{A} + \lambda_1(\mathbf{w}_{T_{1}}^{T}\mathbf{\sigma}_{T_{1}} +  
    \mathbf{w}_{\varphi_{e}}^{T}\mathbf{\sigma}_{\varphi_{e}}) \\ + \lambda_2(\mathbf{w}_{T_{2}}^{T}\mathbf{\sigma}_{T_{2}} + \mathbf{w}_{T_{2}^{*}}^{T}\mathbf{\sigma}_{T_{2}^{*}}),
\label{eq9}
\end{split}
\end{equation}
% \begin{equation}
% \begin{split}
%     \mathcal{L}_{WNNM} = 0.05\mathbf{W}_{A}^{T}\mathbf{\sigma}_{A} + 0.2\mathbf{W}_{T_{1}}^{T}\mathbf{\sigma}_{T_{1}} + \\ 2\mathbf{W}_{T_{2}}^{T}\mathbf{\sigma}_{T_{2}} + 2\mathbf{W}_{T_{2}^{*}}^{T}\mathbf{\sigma}_{T_{2}^{*}} + 0.2\mathbf{W}_{\varphi_{e}}^{T}\mathbf{\sigma}_{\varphi_{e}},
% \label{eq9}
% \end{split}
% \end{equation}

During training, we adopted the mean absolute error loss for the data consistency term. To accommodate GPU memory limitations, we performed the 1D inverse Fourier transform on the acquired k-space data along the fully sampled readout dimension. Subsequently, in each iteration, a small batch of data along the readout direction was selected for training, with a batch size of 4.

\subsection{Experiments Settings}
We compared the proposed method with the LRT reconstruction method \cite{he2016accelerated}, which was implemented using the ADMM algorithm with a total variation regularization term. For temporal basis estimation, we generated an IR-GRE signal dictionary based on the Bloch equations and extracted the $T_1$ basis using SVD. Additionally, $T_2$ and $T_2^*$ bases were extracted from the navigator data. Coil sensitivity maps were derived from the calibration region of an additional GRE scan.

For the retrospective simulation evaluation, we simulated the undersampled data with the following setting: voxel size = 1$\times$1$\times$4 mm$^3$, $\tau$ = 25, 50, 70, 90 ms, $T_E$ = 4.5, 11.2, 17.9, 24.6 ms, $T_R$ = 30 ms, $\alpha$ = 10°, and the number of segments = 80. The simulation data were generated with $R$ = 4 and 8 with a signal-to-noise ratio (SNR) of 30 for evaluation. Different noise levels with SNRs ranging from 10 to 35 were also evaluated on the 4$\times$ undersampled data.

The MRI experiments were conducted on a 3T scanner (uMR790, United Image Healthcare (UIH), Shanghai, China) with a 24-channel head coil. In the phantom study, the SUMMIT imaging protocol parameters were: FOV = 256$\times$256$\times$200 mm$^3$, voxel size = 2$\times$2$\times$5 mm$^3$, $\tau$ = 0, 30, 60, 90, 120 ms, $T_E$ = 3.7, 9.4, 15.1 ms, $T_R$ = 20 ms, $\alpha$ = 4°, number of segments = 128, and scan times = 4.5 minutes for $R$ = 2. We used IR-SE, ME-SE, and ME-GRE sequences as references for quantifying $T_1$, $T_2$, and $T_2^*$ values, respectively. The imaging parameters of these reference sequences are summarized in Table \ref{table1}. 

\begin{table}[ht]
    \begin{center}
      \caption{Imaging protocols of reference scans for MRI experiments.}
      \label{table1}
      \begin{tabular}{ccc}
      \hline
      \multicolumn{3}{c}{\textbf{Phantom Study}}\ \\
      \hline
      \multirow{4}{*}{IR-SE} & \multirow{2}{*}{TI (ms)}   & 100/200/400/700/1000/ \\
                             &                            & 1500/2000/2500/3000 \\
                             & TR (ms)                    & 10000 \\
                             & Scanning time (min)        & 114.0 \\
      \hline
      \multirow{4}{*}{ME-SE} & \multirow{2}{*}{TE (ms)}   & 16.7/33.4/50.2/66.8/ \\
                             &                            & 83.6/100.3/117.0/133.8 \\
                             & TR (ms)                    & 4816 \\
                             & Scanning time (min)        & 8.3 \\
      \hline
      \multirow{3}{*}{ME-GRE} & TE (ms)             & 4.5/11.2/17.9/24.6 \\
                              & TR (ms)             & 30 \\
                              & Scanning time (min) & 1.5 \\
      \hline
      \end{tabular}
    \end{center}
\end{table}

\section{Results}
\label{sec:Results}
\subsection{Retrospective Simulation Evaluation}
\begin{figure}[ht]
    \centering
    \includegraphics[width=80mm]{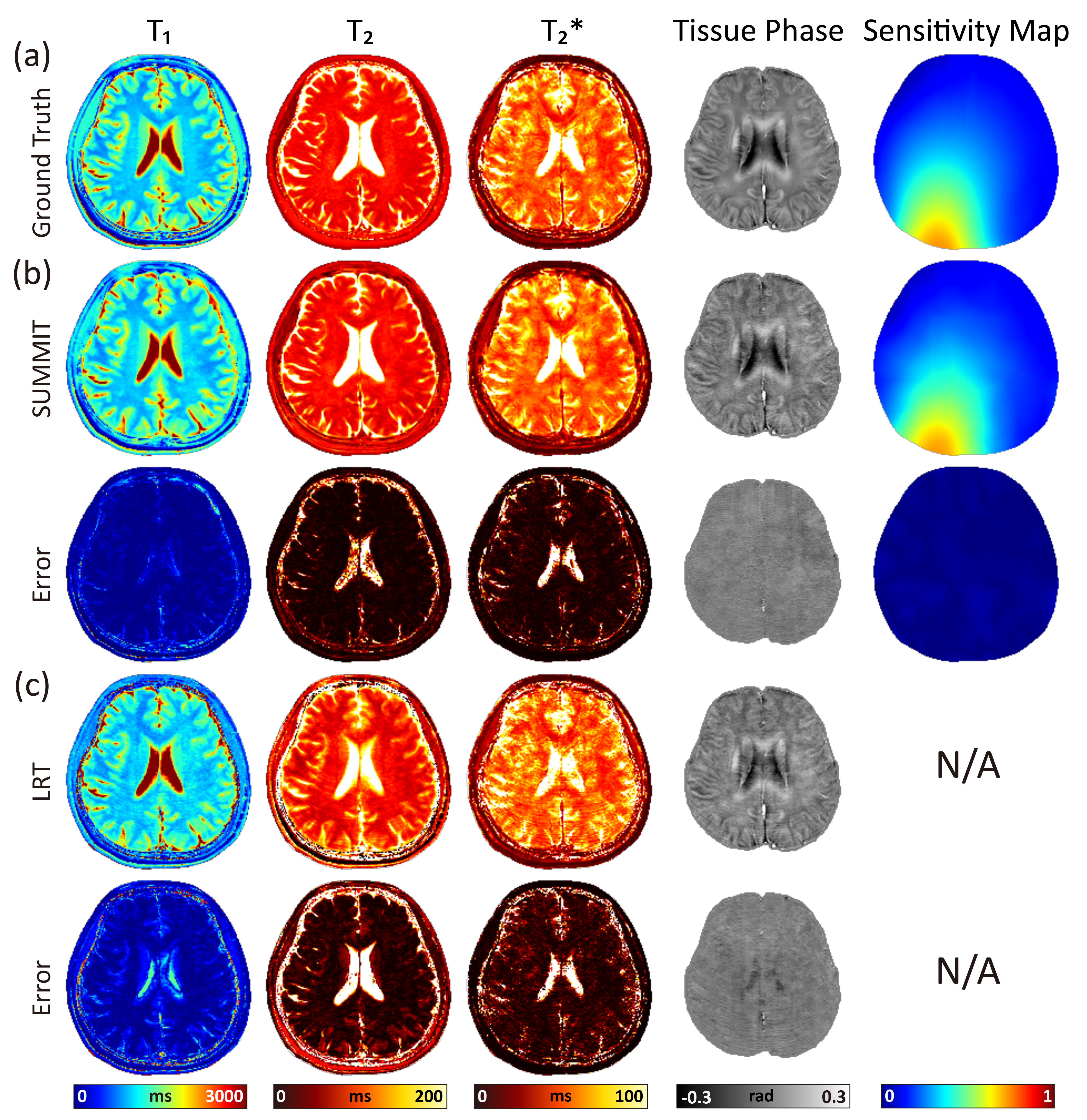}
    \caption{
    Comparison between SUMMIT and LRT on the 4$\times$ retrospective simulation. (a) The ground truth maps used for simulation. (b-c) The reconstructed quantitative maps and corresponding errors of  SUMMIT and LRT.
    }
    \label{fig2}
\end{figure}

\begin{figure}[ht]
    \centering
    \includegraphics[width=72mm]{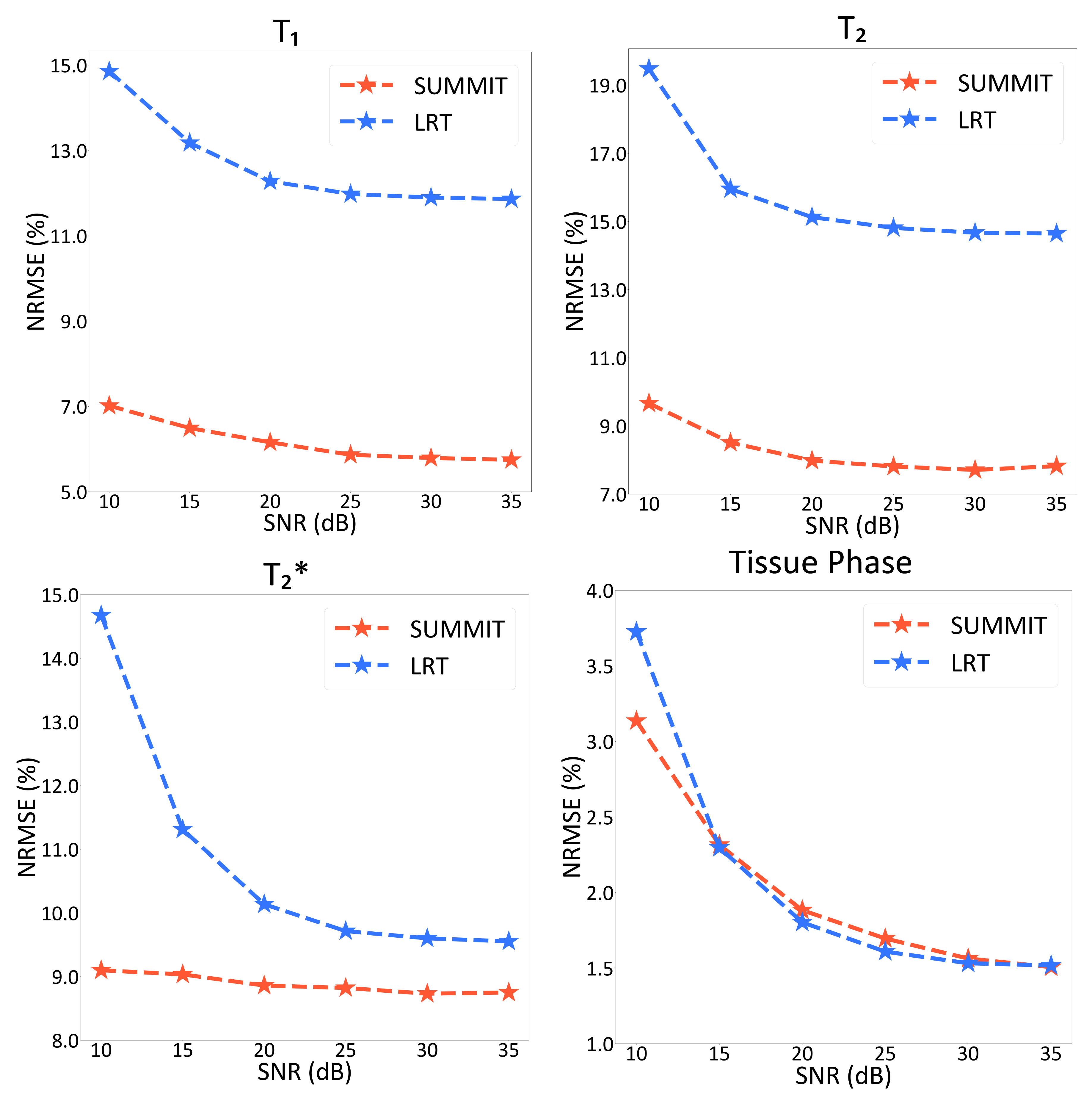}
    \caption{
    The performance variation of SUMMIT and LRT on the 4$\times$ retrospective simulation at different SNRs. SUMMIT shows lower NRMSE on $T_1$, $T_2$, and $T_2^*$ maps compared with LRT.
    }
    \label{fig3}
\end{figure}

\begin{figure}[ht]
    \centering
    \includegraphics[width=80mm]{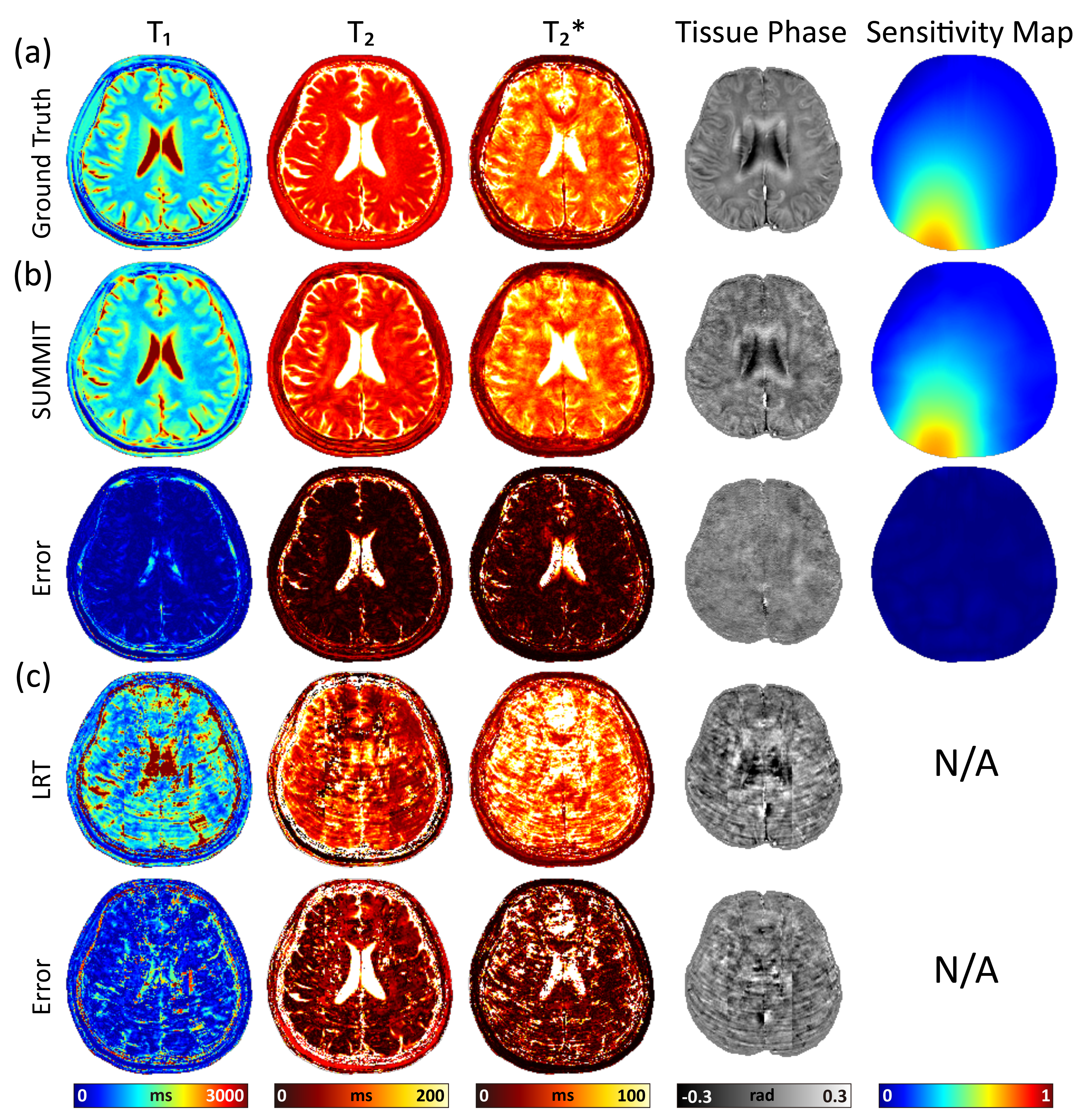}
    \caption{
    Comparison between SUMMIT and LRT on the 8$\times$ retrospective simulation. (a) The ground truth maps used for simulation. (b-c) The reconstructed quantitative maps and corresponding errors of  SUMMIT and LRT.
    }
    \label{fig4}
\end{figure}

Simulation data were obtained using ground truth maps. The reconstructed images and corresponding errors for SUMMIT and LRT are shown in Fig. \ref{fig2}. Compared to LRT, SUMMIT provides superior image quality and reduced reconstruction error for $T_1$, $T_2$, and $T_2^*$ maps. Additionally, the sensitivity maps estimated by SUMMIT closely match the ground truth, whereas LRT requires the ground truth sensitivity maps for reconstruction. Fig. \ref{fig3} displays the normalized root-mean-square error (NRMSE) on $T_1$, $T_2$, $T_2^*$, and tissue phase maps at different SNRs. SUMMIT shows lower NRMSE on $T_1$, $T_2$, and $T_2^*$ maps and similar NRMSE on tissue phase maps across SNRs. SUMMIT's performance remains stable at a low SNR of 10, showing decreased NRMSE by 7.8\%, 9.8\%, and 5.6\% on $T_1$, $T_2$, and $T_2^*$ maps compared to LRT. In addition, the reconstruction error of LRT significantly increases with higher acceleration factors, while SUMMIT maintains lower errors, as indicated by NRMSE reductions of 11.0\%, 11.8\%, and 4.8\% on $T_1$, $T_2$, and $T_2^*$ maps at an extreme $R$ = 8 (Fig. \ref{fig4}).

\subsection{Quantitative Phantom Study}
\begin{figure*}[ht]
    \centering
    \includegraphics[width=181mm]{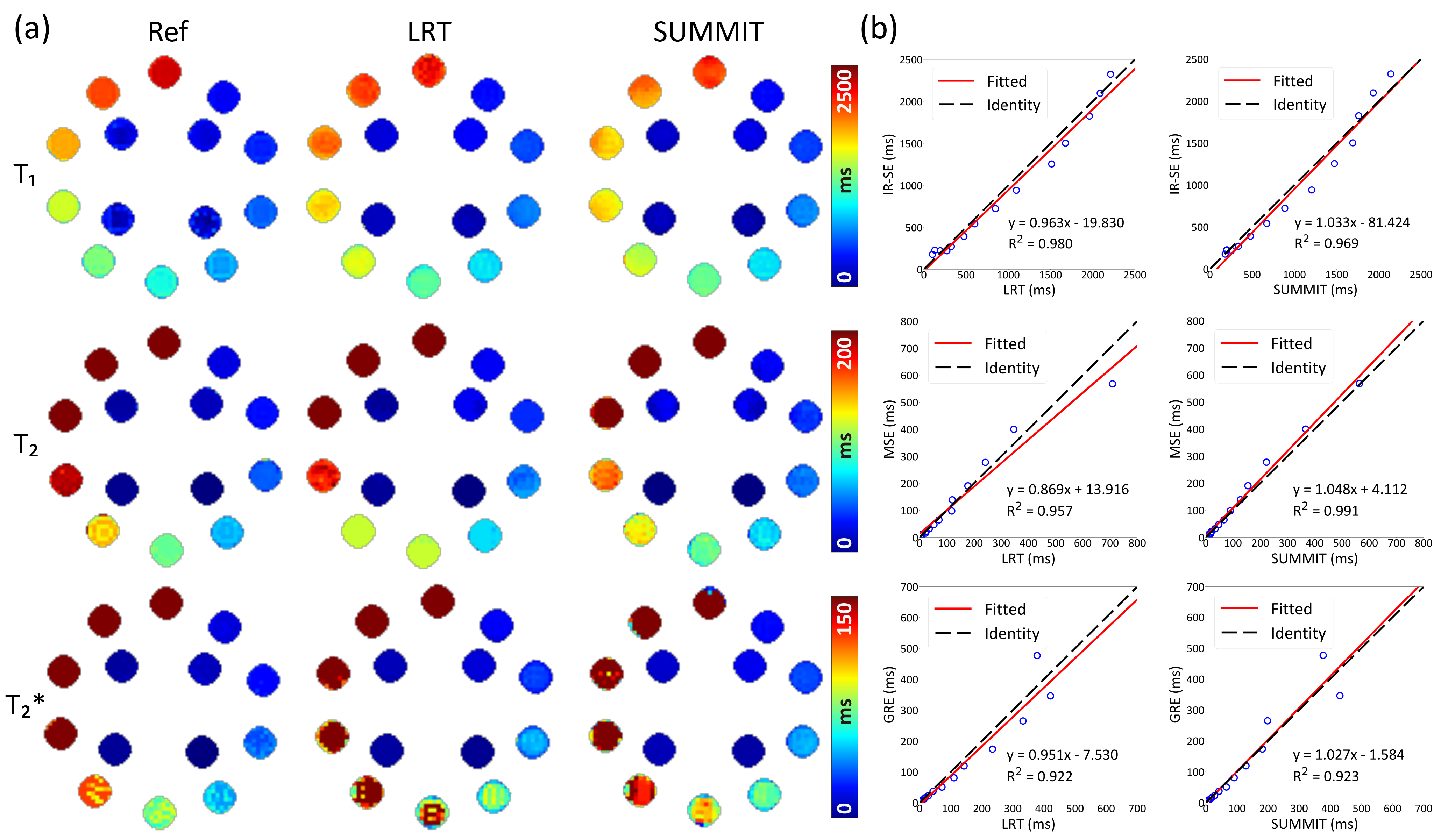}
    \caption{
    Quantitative evaluation on a standard phantom. (a) SUMMIT shows good image quality on the $T_1$, $T_2$, and $T_2^*$ maps, and agrees well with the references (Ref). Those maps from LRT show non-uniformity on the phantom due to the potential reconstruction errors. (b) The correlation analysis indicates SUMMIT correlates well with the reference sequences, as denoted by the coefficient of determination and slope approaching 1. The red line represents the linear regression fitting and the black dashed line corresponds to y=x.
    }
    \label{fig5}
\end{figure*}

To evaluate the efficiency of SUMMIT in real-world scenarios, we conducted a prospective study using a standard ISMRM/NIST phantom \cite{stupic2021standard}. Fig. \ref{fig5}(a) presents the reconstruction results from prospectively undersampled data with an acceleration factor $R$ = 2, along with correlation analysis in Fig. \ref{fig5}(b). The $T_1$, $T_2$, and $T_2^*$ values estimated by SUMMIT and reference sequences demonstrate good agreement, as indicated by the correlation slope approaching 1 with $R^2$ = 0.969, 0.991, and 0.923 for $T_1$, $T_2$, and $T_2^*$ respectively. In contrast, the LRT consistently underestimates $T_2$ values (slope = 0.869 for LRT \textit{vs.} slope = 1.048 for SUMMIT), likely due to errors in the estimation of temporal basis. Both LRT and SUMMIT exhibit deviations from reference $T_2^*$ values when $T_2^*$ ranges exceed 200 ms, which might be attributable to inherent limitations of the imaging parameters.

\section{Discussion}
\label{sec:Discussion}
In this work, we proposed SUMMIT, an unsupervised approach for 3D simultaneous MP-qMRI mapping using INR. SUMMIT aims to obtain $T_1$, $T_2$, $T_2^*$, and QSM images from a single MRI scan. Unlike the conventional LRT method, SUMMIT solves a single-step parameter optimization problem, avoiding potential error propagation from multiple reconstruction steps, such as temporal basis estimation, high-dimensional tensor reconstruction, and parameter fitting. SUMMIT does not rely on additional calibration data for coil sensitivity estimation. We validated SUMMIT using simulation data, a prospective phantom study. The simulation results demonstrate its superiority at lower SNRs and higher acceleration factors with less reconstruction error and fewer artifacts. The quantitative phantom study further highlights the good quantitative agreement with reference images. However, deviations are noted in $T_1$ and $T_2$ maps, particularly in cerebrospinal fluid regions with high $T_1$ and $T_2$ values. These discrepancies arise from inherent limitations in imaging parameters, which introduce biases in relaxation processes and lead to underestimation of $T_1$ and $T_2$ values. ME-SE quantifies $T_2$ values by employing a series of selective 180° refocusing pulses to obtain multi-echo data within a single TR. Imperfections in the 180° pulse may generate additional echoes, potentially contaminating the signal decay \cite{lebel2010transverse}.

Future work will further improve our method by addressing several current limitations. Firstly, the current framework lacks robustness against motion artifacts. The total scan time of our sequence is slightly longer than that of typical clinical scans, increasing susceptibility to motion during imaging. By incorporating navigator modules, the sequence would enable motion detection and capture, allowing us to resolve motion states and generate motion-robust quantitative maps either by modeling motion states or correcting motion-corrupted data during reconstruction. Secondly, the current reconstruction framework underutilizes potential correlation information between different quantitative maps, such as the inherent structural consistency in naturally aligned images. Future research aims to explore more effective ways to leverage this shared information. This may involve developing a more streamlined representation strategy that distinguishes common and unique information across different quantitative maps, ultimately enhancing the stability and accuracy of MP-qMRI reconstruction. Thirdly, the INR-based reconstruction takes approximately one hour due to multiple steps of high-dimensional tensor computations and gradient descent, similar to existing LRT-based reconstruction techniques. Meta-learning shows promise in learning the initial network weights as strong priors \cite{tancik2021learned}, thus enabling faster convergence during optimization and improving clinical efficiency. In summary, our future work will focus on improving motion robustness, exploiting correlation information between quantitative maps more effectively, and optimizing reconstruction times to advance the clinical utility of our MP-qMRI method.

\section{Conclusion}
\label{sec:Conclusion}
In this study, we developed a novel 3D multiparametric qMRI sequence and a zero-shot reconstruction strategy via coordinate-based neural representation. The proposed imaging methodology encodes $T_1$, $T_2$, $T_2^*$, and phase maps into a high-dimensional k-space that is undersampled along the spatiotemporal dimensions. Then, we adopted the concept of neural representation to model a series of unknown variables as functions of spatial coordinates, enabling the direct reconstruction from k-space to quantitative maps, thereby eliminating the error propagation. The experimental results of numerical simulation and phantom data demonstrate the advantages of the proposed direct reconstruction strategy in enhancing quantitative accuracy. 

\bibliographystyle{IEEEtran}
\balance
\bibliography{ref}
\end{document}